\documentclass{AVEC_EA}
\usepackage{amsmath}
\usepackage{supertabular} 

\begin{document}
\twocolumn[
\begin{titlepage}
    \papertitle{Racing a Wheeled Quadruped: Active Load Transfer Mitigation via Model Predictive Control}
    \authors{Marla Eisman\textsuperscript{1)}~~~Brian Lam\textsuperscript{1)}~~~Samuel Sonnino\textsuperscript{2)}~~~Francesco Borrelli\textsuperscript{1)}
    }
    \contactinfo{1) University of California, Berkeley \\
    2) Politecnico di Milano, Italy \\
    (Corresponding author: {\normalfont marla\_eisman@berkeley.edu})
    }
    \myabstract{ This paper presents a hierarchical control framework using model predictive control (MPC) and reinforcement learning (RL) for active roll control to manage lateral load transfer during autonomous racing of a wheeled quadruped. The framework integrates offline time-optimal raceline generation, an online MPC planner that actively minimizes the lateral Load Transfer Ratio (LTR), and a low-level, whole-body RL policy deployed directly onto the robot's 16 actuators. The MPC is based on a vehicle dynamics bicycle model of the Unitree Go2-W platform. The robot's leg actuators act as active suspension where knee joints generate anti-roll torque to bank into turns. Physical track experiments demonstrate that active roll control reduces mean LTR by up to 44\%, improves the fastest lap time by 8.7\%, and boosts peak lateral acceleration capability by 21.3\% to 1.98 m/s$^2$, maintaining robust high-speed stability beyond the range of a non-tilting baseline controller. Supplementary code and video can be found at \url{https://github.com/meisman-ucb/go2w-roll-control-mpc}.}
\end{titlepage}
]
\thispagestyle{fancy}

\section{Introduction} 
Wheeled quadrupeds combine the flexibility of legged robotics with the efficiency and speed of wheeled vehicles.  Existing  control designs in the robotics field apply legged, gait-based locomotion dynamics to wheeled quadruped control~\cite{survey}. This paper approaches the problem from a vehicle dynamics perspective with a focus on differential torquing, rollover risk and load transfer effects. 

\begin{figure}
    \centering
    \includegraphics[width=0.6\linewidth]{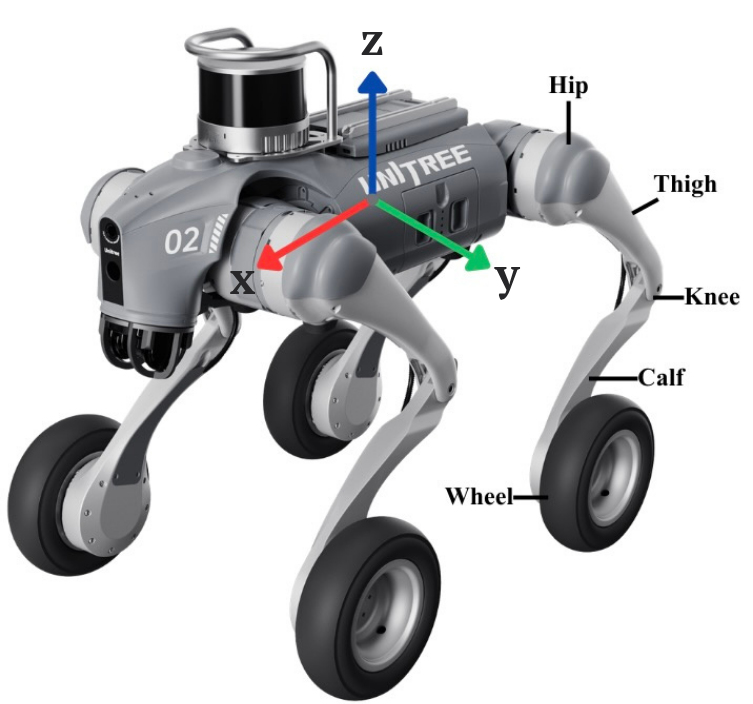}
    \caption{Unitree Go2-W robot platform.}
    \label{fig:go2w}
\end{figure}

\begin{figure}
    \centering
    \includegraphics[width=0.8\linewidth]{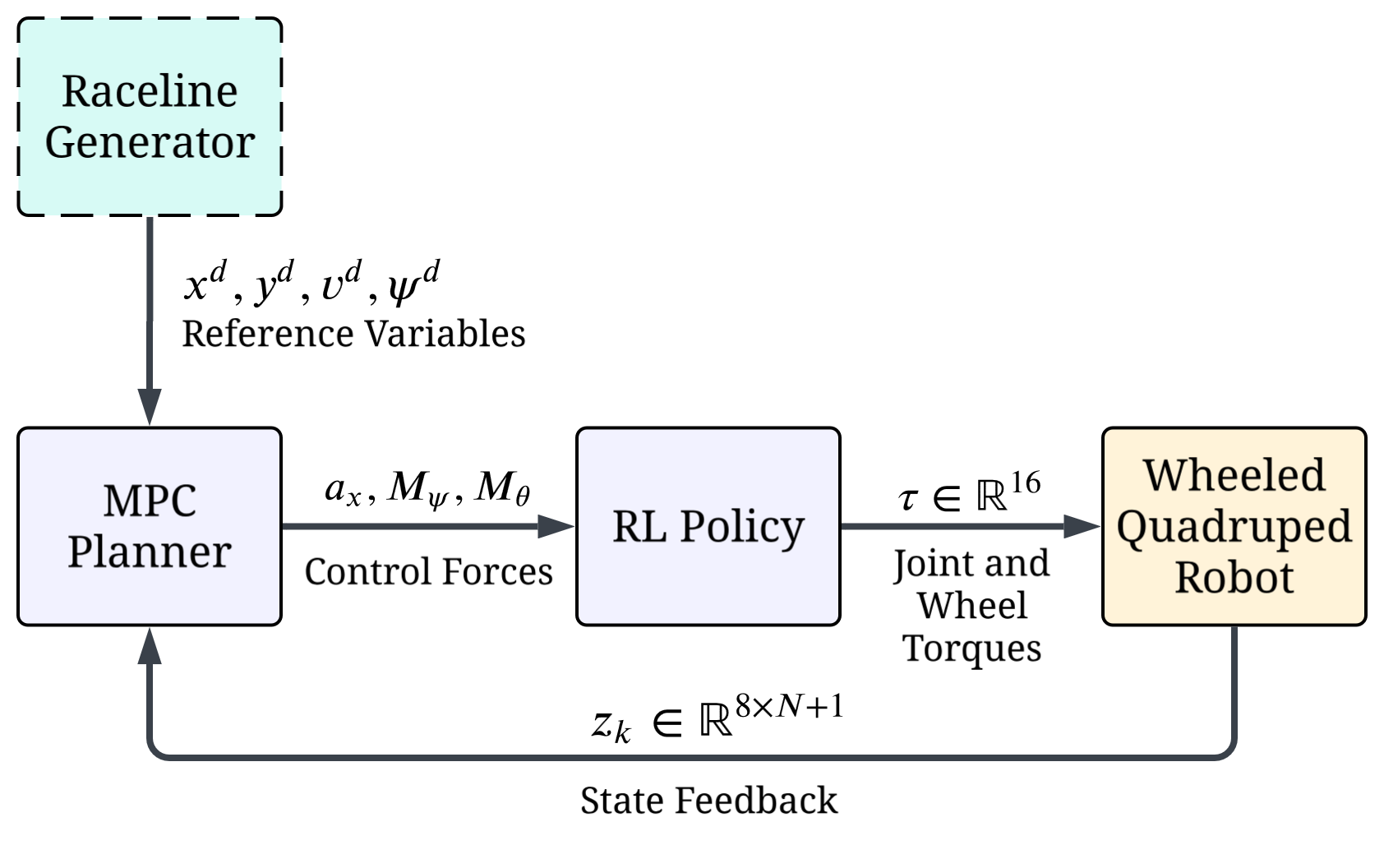}
    \caption{Hierarchical control system.}
    \label{fig:diag}
\end{figure}

The control of a wheeled quadruped is challenging due to the complex contact dynamics and 16 degrees of freedom (12 joints + 4 wheels). This paper presents a model predictive control (MPC) formulation for racing lap-time minimization by modeling the Unitree Go2-W (Fig.~\ref{fig:go2w}) as a dynamic bicycle with active roll control. A nominally symmetric stance is held stiffly to approximate a rigid vehicle chassis. The MPC model exploits the knee joints to generate roll torque to induce active roll control. The knee joints bending asymmetrically (left versus right) generates anti-roll torque during cornering to reduce lateral load transfer. When the roll dynamics are not modeled, the MPC treats the vehicle as a planar rigid body. The dynamics are integrated into a control architecture including (1) an offline dynamics-based raceline optimizer, (2) online MPC for reference tracking, and (3) an online low-level reinforcement learning (RL) policy to command joints to achieve the MPC outputs (Fig.\ref{fig:diag}). We demonstrate experimentally that the active roll control of this approach yields higher stability and racing performance compared to a non-tilting baseline. 

This paper is structured as follows: Section~\ref{sec:background} discusses related work and specific contributions of this paper. Section~\ref{sec:dyn} introduces the vehicle model and load transfer ratio (LTR) metric. Section ~\ref{sec:raceline} explains the optimal raceline generation. Section~\ref{sec:control} describes the MPC formulation, cost function, and constraints. Section~\ref{sec:rl} illustrates the simulated RL training and deployment of the policy in the control architecture. Sections~\ref{sec:exp} and~\ref{sec:results} discuss the experimental setup for validation and results. All constant parameters are tabulated in Sec.~\ref{sec:appendix}.
\section{Background}\label{sec:background}
Autonomous racing pushes vehicles to the edge of their dynamic capabilities, where accurate models and online optimization are essential for lap-time minimization. MPC has become a standard tool for this goal, with learning-based extensions reducing reliance on perfect models~\cite{lmpc} and online adaptation handling parameter drift~\cite{costa}. These formulations, however, assume a rigid vehicle with fixed body geometry, leaving no flexibility to exploit the additional degrees of freedom that wheel-legged platforms inherently provide. 

Wheeled quadrupedal robots combine the speed and efficiency of wheeled vehicles with the terrain adaptability of legged systems, but utilizing both benefits requires whole-body coordination across highly coupled dynamics. Whole-body motion planning and control was introduced for wheeled quadrupeds on the ANYmal platform~\cite{bjel} and later extended to complex terrain such as stair climbing~\cite{bjel2}. In parallel, work on actuated quadruped spines has shown that traditionally rigid body segments can serve as performance-enhancing control inputs rather than fixed constraints~\cite{spine}. These efforts target terrain traversal and gait stability, however, and do not yet exploit the high-speed dynamic performance that wheels enable.

At high lateral accelerations, load transfer reduces the effective cornering capacity of the outer tires and raises rollover risk. Counteracting load transfer by leaning the body inward is a well-established strategy in road vehicles, high-speed trains, and motorcycles. Narrow tilting vehicles achieve this through dedicated tilt hardware on a specialized chassis~\cite{liang}, an approach that does not transfer to the compact hardware of robotic platforms. Recent work has extended active roll concepts to wheel-legged systems. MPC has been used to regulate roll on rough terrain on a knee-wheeled robot~\cite{pan}, and a linear quadratic regulator with sliding-mode control has been proposed for rollover avoidance during skid-steering maneuvers on a wheel-legged vehicle~\cite{liu}. However, both are validated in simulation only, and neither targets a racing objective. 

To our knowledge, no prior work demonstrates active roll control on a wheel-legged quadruped for a racing objective, nor experimentally validates rollover avoidance for skid-steering wheeled quadrupeds. These platforms must generate yaw entirely through differential wheel torque and are especially susceptible to load transfer at high yaw rates. This work addresses both gaps. 

The primary contributions of this paper are: 
\begin{enumerate}
    \item MPC formulation with active load transfer mitigation via roll control to enhance racing performance of a wheeled quadruped. 
    \item Hierarchical controller including offline racing line optimization, online MPC for reference tracking, and joint-level RL control.
    \item Successful hardware validation of proposed method.
\end{enumerate}

\begin{figure}
    \centering
    \includegraphics[width=0.8\linewidth]{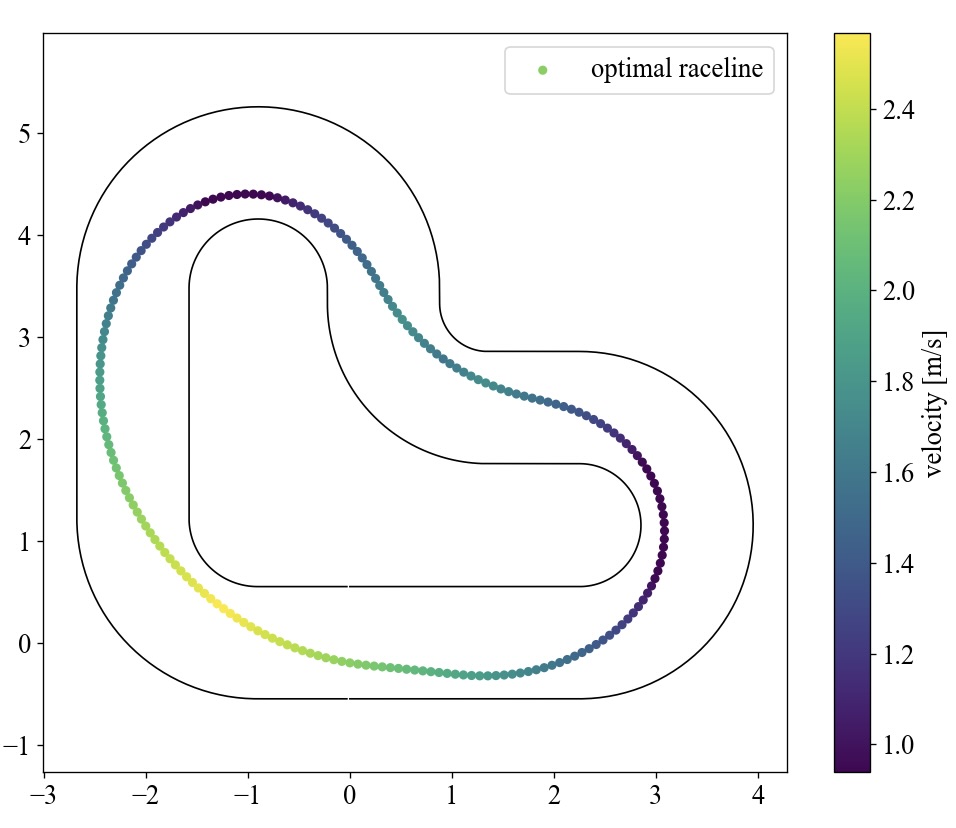}
    \caption{Optimal raceline with velocity.}
    \label{fig:raceline}
\end{figure}

\section{Vehicle Dynamics Model}\label{sec:dyn}

We model the wheeled quadruped as a dynamic bicycle with an added active roll degree of freedom. The bicycle representation lumps the four wheels into an equivalent front wheel and rear wheel, capturing the dominant cornering dynamics while remaining efficient enough to solve in real time. The Go2-W has no steered wheels so yaw is generated by differential wheel torque (skid-steering) and body roll is generated by differential leg vertical extension acting as adjustable suspension. The conventional steering input is therefore replaced by two moment inputs: a yaw moment $M_\psi$ and a roll moment $M_\theta$ alongside longitudinal acceleration $a_x$ for forward motion. 

The model relies on three assumptions:
\begin{enumerate}
\item \label{asm:1}
The robot is modeled as a rigid body with constant mass, center of gravity (CG), and moments of inertia $I_x$, $I_z$. All joints except the active roll degree of freedom are held in a nominal standing pose, so the body geometry (track width $w$ and wheelbase $L_f+L_r$) is constant.
\item \label{asm:2}
The vehicle's lateral and yaw dynamics are captured by a bicycle model~\cite{rigid} with sideslip $\beta$ and yaw rate $\dot\psi$, where front and rear tire forces are linear functions of slip angle with constant cornering stiffnesses $C_{\alpha f}$ and $C_{\alpha r}$.
\item \label{asm:3}
All four wheels remain in contact with the ground at all times.
\end{enumerate}

The nominal standing pose in Assumption~\ref{asm:1} is enforced in practice by the low-level RL policy (Sec.~\ref{sec:rl}), which is trained to maintain joint symmetry and a fixed body configuration apart from the commanded roll. The linear tire model in Assumption~\ref{asm:2} is valid in the moderate-slip regime encountered at our target speeds and is preserved by the active roll controller itself, which keeps the lateral load distributed symmetrically across the left and right wheels. Continuous ground contact (Assumption~\ref{asm:3}) is similarly maintained by minimizing load transfer.

\subsection{State and Control Notation} \label{sec:model}
Assumptions~\ref{asm:1}--\ref{asm:3} yield vehicle dynamics $\dot z = f(z,u)$ with state
\begin{equation} \label{eq:state}
z=\begin{bmatrix}x & y & \psi & \dot{\psi} & \theta & \dot{\theta}& v&\beta\end{bmatrix}^\top\in \mathbb{R}^8,
\end{equation}
where $(x,y)$ are the 2D global position coordinates of the CG, $\psi$ and $\dot\psi$ are yaw angle and yaw rate, $\theta$ and $\dot\theta$ are body roll angle and roll rate, $v$ is the longitudinal speed, and $\beta$ is the body sideslip angle. The vertical position is not tracked because the robot operates on a planar surface and the CG height $h$ is treated as constant under Assumption~\ref{asm:1}. The control input is
\begin{equation}
u=\begin{bmatrix} a_x & M_\psi & M_\theta \end{bmatrix}^\top\in \mathbb{R}^3,
\end{equation}
where $a_x$ is longitudinal acceleration, $M_\psi$ is the yaw moment, and $M_\theta$ is the roll moment.

\subsection{Planar Dynamics}
The 2D position and longitudinal speed are expressed as~\cite{rigid}
\begin{equation}
\dot x = v\cos(\psi+\beta), \quad \dot y = v\sin(\psi+\beta), \quad \dot v = a_x.
\label{eq:kin}
\end{equation}
Movement along the track produces $\psi$, $\beta$, and $v$ which provide direction and speed. The first two terms are computed as~\cite{rigid}
\begin{align}
\dot\beta &= \frac{F_{y,f}+F_{y,r}}{mv}-\dot\psi, \label{eq:beta_force}\\
\ddot\psi &= \frac{L_f F_{y,f}-L_r F_{y,r}+M_\psi}{I_z}, \label{eq:psi_force}
\end{align}
where $F_{y,f}$ and $F_{y,r}$ are the lumped front and rear lateral tire forces, $L_f$ and $L_r$ are the longitudinal distances from the CG to the front and rear axles, and $I_z$ is the moment of inertia about the z-axis.

\subsection{Tire Model}
Using small-angle approximations and neglecting lateral velocity, the front and rear slip angles are~\cite{rigid}
\begin{equation}
\alpha_f=\beta+\frac{L_f\dot\psi}{v},\quad \alpha_r=\beta-\frac{L_r\dot\psi}{v},\quad v>0,
\label{eq:slip}
\end{equation}
and the linear tire model expresses lateral force as:
\begin{equation}
F_{y,f}=C_{\alpha f}\alpha_f,\quad F_{y,r}=C_{\alpha r}\alpha_r,
\label{eq:tireforce}
\end{equation}
with constant cornering stiffnesses $C_{\alpha f}$, $C_{\alpha r}$ (Table~\ref{tab:params}). Substituting Eqns.~\eqref{eq:slip}--\eqref{eq:tireforce} into
Eqns.~\eqref{eq:beta_force}--\eqref{eq:psi_force} yields the closed-form sideslip and yaw dynamics:
\begin{align}
\dot\beta &= -\frac{C_{\alpha f}+C_{\alpha r}}{mv}\beta+\left(\frac{-C_{\alpha f}L_f+C_{\alpha r}L_r}{mv^2}-1\right)\dot\psi, \label{eq:beta}\\
\ddot\psi &= \frac{-C_{\alpha f}L_f+C_{\alpha r}L_r}{I_z}\beta - \frac{C_{\alpha f}L_f^2+C_{\alpha r}L_r^2}{I_z v}\dot\psi + \frac{M_\psi}{I_z}. \label{eq:psi}
\end{align}
Together with \eqref{eq:kin}, equations \eqref{eq:beta}--\eqref{eq:psi} fully describe the planar motion in $(x,y,\psi,\dot\psi,v,\beta)$ given the inputs $a_x$ and $M_\psi$.

\subsection{Roll Dynamics and Load Transfer} \label{sec:ltr}
The remaining state, body roll $\theta$, is decoupled from the planar dynamics, but $\theta$ is driven by the lateral acceleration produced by the planar motion. Roll moments about the longitudinal axis give
\begin{equation}
\ddot\theta = \frac{1}{I_x}\bigl(-mgh\sin\theta - m a_y h\cos\theta + M_\theta\bigr),
\label{eq:roll}
\end{equation}
where $a_y = v(\dot\beta+\dot\psi)$ is the lateral acceleration of the CG, $g$ is gravitational acceleration, $h$ is the CG height, and the $-mgh\sin\theta$ term is the gravitational restoring torque.

In cornering, $a_y$ shifts the load from the inner to the outer wheels. We quantify this with the load transfer ratio (LTR), computed from a separate left-right two-wheel projection of the robot (distinct from the front-rear bicycle projection used for the planar dynamics):
\begin{equation}
\text{LTR}=\frac{2h}{w}\left(\frac{a_y}{g}\cos\theta+\sin\theta\right),
\label{eq:ltr}
\end{equation}
where $w$ is the track width. $\text{LTR}=0$ corresponds to equal vertical force on the left and right wheels, while $|\text{LTR}|=1$ corresponds to wheel lift-off and imminent rollover.

Driving LTR toward zero prevents rollover by keeping all four wheels loaded and in contact and preserves the symmetric lateral loading under which the constant cornering stiffnesses $C_{\alpha f}$, $C_{\alpha r}$ remain valid, which justifies decoupling the roll dynamics in \eqref{eq:roll} from the planar dynamics in \eqref{eq:beta}--\eqref{eq:psi}.

Setting $\text{LTR}=0$ in \eqref{eq:ltr}, applying a small-angle approximation in $\theta$, and assuming $|\dot\beta|\ll|\dot\psi|$ yields the optimal roll angle:
\begin{equation}\label{eq:thetastar}
\theta^*=-\arctan\left(\frac{v\dot\psi}{g}\right)\approx -\frac{v\dot\psi}{g}.
\end{equation}
This is the banking angle at which gravity exactly cancels the centripetal contribution to lateral load transfer. 

The continuous dynamics (Eqns.~\eqref{eq:kin},~\eqref{eq:beta},~\eqref{eq:psi}, and~\eqref{eq:roll}) are discretized and integrated using the fourth-order Runge-Kutta method~\cite{ode}, denoted by the discrete dynamics $z_{k+1} = f_d(z_k,u_k)$.

\section{Offline Raceline Generation}\label{sec:raceline}
Given a track, the optimal raceline is computed by solving a FTOCP that minimizes lap time subject to the track boundaries and the dynamics $\dot{z} = f(z,u)$ detailed in Sec.~\ref{sec:dyn}, using the approach in~\cite{raceline}. An example of an optimal raceline generated for the experimental track is shown in Fig.~\ref{fig:raceline}. The raceline is stored as
\begin{equation}
z^d(s) = [\,x^d(s),\; y^d(s),\; \psi^d(s),\; v^d(s)\,]^\top,
\label{eq:zd}
\end{equation}
where $s$ is the arc-length coordinate along the centerline and $z^d$ contains the desired position, heading, and velocity components of the full state vector $z$ described in Eqn.~\eqref{eq:state}. In the online MPC discussed next, we use the optimal raceline from Eqn.~\eqref{eq:zd} as the reference.

\section{Online MPC}\label{sec:control}

The offline-computed raceline is tracked with an MPC. At each time step, the current robot state is measured, the nearest point on the raceline is found, and a finite-time optimal control problem (FTOCP) is solved over a prediction horizon of $N$ steps. The first input of the optimal control sequence is applied to the robot and the procedure repeats at the next time step~\cite{mpc}.

\subsection{Optimal Control Problem}\label{sec:ocp}
At time step $k$, let $z_k$ denote the measured robot state and $z_{k+i|k}$ denote the state predicted for time $k+i$ from the information available at time $k$. Using the same convention, we denote control inputs over the prediction horizon as $u_{k+i|k}$ for $i \in \{0, \dots, N-1\}$. The prediction sequences over the horizon $N$ are 
\begin{align*}
Z_k = &[\,z_{k|k},\; z_{k+1|k},\; \dots,\; z_{k+N|k}\,], \\
U_k = &[\,u_{k|k},\; u_{k+1|k},\; \dots,\; u_{k+N-1|k}\,].
\end{align*}
The optimization is performed over $Z_k$ and $U_k$, which are coupled through the system dynamics constraint and initial state. The raceline reference is denoted $Z^d_k = [\,z^d_{k|k},\; \dots,\; z^d_{k+N|k}\,]$. Each reference state $z^d_{k+i|k}$ is obtained by evaluating the raceline~\eqref{eq:zd} at arc-length coordinates determined by the reference velocity profile and time step. The FTOCP solved at each time step is 

\begin{subequations}\label{eq:ftocp}
\begin{align}
J^\star(z_k) = & \min_{U_k} \quad \sum_{i=0}^{N} w^\top \ell\bigl(z_{k+i|k},\,z^d_{k+i|k}\bigr) \nonumber \\
& + \sum_{i=0}^{N-1} u_{k+i|k}^\top R\, u_{k+i|k}
 + \sum_{i=1}^{N-1} \Delta u_{k+i|k}^\top R_\Delta\, \Delta u_{k+i|k} \label{eq:ftocp_cost}\\
\text{s.t.}\quad
& z_{k|k} = z_k, \label{eq:ftocp_ic}\\
& z_{k+i+1|k} = f_d\bigl(z_{k+i|k},\, u_{k+i|k}\bigr), \forall i \in \{0,\dots,N-1\}, \label{eq:ftocp_dyn}\\
& v_{\min} \le v_{k+i|k} \le v_{\max}, \forall i \in \{0,\dots,N\}, \label{eq:ftocp_vbound}\\
& a_{x,\min} \le a_{x,k+i|k} \le a_{x,\max}, \forall i \in \{0,\dots,N-1\}, \label{eq:ftocp_abound}\\
& |M_{\theta,k+i|k}| \le M_{\theta,\max}, \forall i \in \{0,\dots,N-1\}, \label{eq:ftocp_Mbound}
\end{align}
\end{subequations}
where $\Delta u_{k+i|k} = u_{k+i|k} - u_{k+i-1|k}$, $R$ is the control weight matrix, $R_\Delta$ is the control-rate weight matrix, $w$ is the state-tracking weight vector, and $\ell(\cdot,\cdot)$ is the vector of state-tracking cost terms. Equation \eqref{eq:ftocp_ic} is the initial condition, \eqref{eq:ftocp_dyn} is the discretized dynamics from Sec.~\ref{sec:dyn}, and \eqref{eq:ftocp_vbound}--\eqref{eq:ftocp_Mbound} are the hard input and state constraints. This is a tracking formulation~\cite{mpc} with no terminal cost or constraint.

At each time step $k$, the FTOCP~\eqref{eq:ftocp} is solved obtaining the optimal sequence $U_k^\star$, and only the first element $u_{k|k}^\star$ is commanded to the robot. The horizon advances one step, and the problem is re-solved with the updated initial condition. 

\subsection{Stage Cost}\label{sec:cost}
% The cost function is defined as \begin{equation}
% J_N(z_{0,k}, U_k)=\sum^N_{i=0}q(i,z_i)+\sum^{N-1}_{i=0}u_i^\top Ru_i+\sum^{N-1}_{i=1}u_\Delta^\top R_\Delta u_\Delta
% \label{eq:cost}
% \end{equation} where $u_\Delta =u_i-u_{i-1}$, $R$ is the control weight matrix, and $R_\Delta$ is the control rate weight matrix. The stage cost $q(i,z_i)$ includes 8 state-dependent penalty terms defined below.

% \subsubsection{Stage Cost}\label{sec:stage}
The state-based component of the stage cost is a weighted sum $w^\top \ell(z_i, z^d_i)$ where the state-tracking cost function $\ell(\cdot, \cdot)$ is defined as follows, omitting the subscript $k+i|k$ for brevity: 
\begin{equation}\label{eq:stagecost} 
\ell(z_i, z^d_i) =
\begin{bmatrix}
1-\cos(\psi_i-\psi^d_i) \\
(v_i-v^d_i)^2 \\
(x_i-x^d_i)^2 + (y_i-y^d_i)^2 \\
\max(0,\beta_i-\beta_{\max})^2 + \max(0,-\beta_i-\beta_{\max})^2 \\
\max(0,\dot\psi_i-\dot\psi_{\max})^2 + \max(0,-\dot\psi_i-\dot\psi_{\max})^2 \\
\bigl[\max\bigl(0,\;-\sin(\psi^d_i)(x_i-x^d_i)+\cos(\psi^d_i)(y_i-y^d_i)\bigr)\bigr]^2 \\
(v_i\dot\psi_i + g\,\theta_i)^2 \\
\max(0,\theta_i-\theta_{\max})^2 + \max(0,-\theta_i-\theta_{\max})^2
\end{bmatrix}
\end{equation}
with weight vector $w = [\,w_\psi,\, w_v,\, w_p,\, w_\beta,\, w_{\dot\psi},\, w_{tb},\, w_{\theta^\star},\, w_\theta\,]^\top$. The terms, in order, serve the following roles: The $1-\cos(\psi_i-\psi^d_i)$ term tracks the heading reference smoothly while avoiding angle-wrapping discontinuities.
 The $(v_i-v^d_i)^2$ term tracks the reference speed.
 The $(x_i-x^d_i)^2 + (y_i-y^d_i)^2$ term tracks the reference position to keep the trajectory near the raceline. 
 The $\beta$ penalty limits sideslip outside of $[-\beta_{\max}, \beta_{\max}]$, keeping the robot in the linear-tire regime under which the bicycle model (Sec.~\ref{sec:dyn}) is valid.
 The $\dot\psi$ penalty similarly activates outside of $[-\dot\psi_{\max}, \dot\psi_{\max}]$ to limit yaw rate.
 The track-boundary term penalizes lateral deviation from the raceline tangent at $\psi^d_i$. 
 The LTR is driven to zero when $v_i\dot\psi_i + g\,\theta_i =0$, so the term $(v_i\dot\psi_i + g\,\theta_i)^2$ drives the roll angle to the LTR\,$=0$ condition as seen in \eqref{eq:thetastar}.
 Finally, the $\theta$ penalty activates outside $[-\theta_{\max}, \theta_{\max}]$ to prevent commanded banking beyond a safe rollover margin.
 The bounding penalties ($\max(0,\cdot)^2$) act as soft constraints to preserve recursive feasibility while strongly discouraging violations~\cite{mpc}.

\subsection{Constraints}\label{sec:constraints}
The bounds in \eqref{eq:ftocp_vbound}--\eqref{eq:ftocp_Mbound} are chosen from the safe operating conditions for experimentation and nominal motor torque limits of the Go2-W to avoid overcurrent shutdown~\cite{go2w}. The bounds $v_{\min}$ and $v_{\max}$ constrain forward speed, while $a_{x,\min}$ and $a_{x,\max}$ constrain longitudinal acceleration, and
$M_{\theta,\max}$ constrains the roll moment. Numerical values are listed in Table~\ref{tab:params}. The yaw moment $M_\psi$ is not included in the constraints because the wheel torque limits are instead enforced in the RL policy on the low-level layer and penalized through the $\dot\psi$ term in the stage cost \eqref{eq:stagecost}.

\section{Online RL Low-Level Control}\label{sec:rl}

The MPC defined in Sec.~\ref{sec:model} outputs a high-level, $3$-input command $[a_x, M_\psi, M_\theta]^\top$, but the Go2-W is actuated by 16 individual motors. The contact and motor dynamics are highly coupled, making an analytical relationship between high and low-level commands impractical and computationally expensive. Instead, we train an RL policy in simulation with domain randomization. The low-level policy receives the MPC commands as part of its observation and produces joint-level targets that the Go2-W's built-in PD controller converts to 16-dimensional motor torques. 

\subsection{Policy Architecture}
The policy is a multi-layer perceptron (MLP) with hidden dimensions $[512,256,128]$ and ELU activations, mapping a $53$-dimensional observation vector to a 16-dimensional action vector at every control step. The observation includes linear velocity in the body frame, angular velocity in the body frame, the MPC command $[a_x, M_\psi, M_\theta]^\top$, leg joint positions relative to the default standing configuration $q_\ell \in \mathbb{R}^{12}$, all motor velocities $\dot q \in \mathbb{R}^{16}$, and the previous action $a_{t-1} \in \mathbb{R}^{16}$. The action $a_t$ consists of
\begin{equation}
a_t = [\,q_\ell^d,\; \dot q_w^d\,]^\top \in \mathbb{R}^{16},
\end{equation}
with the desired positions of the leg joints $q_\ell^d \in \mathbb{R}^{12}$ and the desired velocities of the wheel joints $\dot q_w^d \in \mathbb{R}^{4}$. The Go2-W's onboard firmware converts each joint $i$ to torques through a PD controller
\begin{equation}
\tau_i = K_{p,i}\bigl(q_i^d - q_i\bigr) + K_{d,i}\bigl(\dot{q}_i^d - \dot{q}_i\bigr),
\end{equation}
where $q_i$ and $\dot{q}_i$ are the measured position and velocity of joint $i$, $K_{p,i}$ and $K_{d,i}$ are joint-specific stiffness and damping gains.

\subsection{Training}
The policy is trained with Proximal Policy Optimization (PPO)~\cite{ppo} and Generalized Advantage Estimation (GAE)~\cite{gae} in Isaac Lab~\cite{isaac, robot} across $4096$ parallel environments. The reward function encourages tracking of the body-level commands $[a_x, M_\psi, M_\theta]^\top$ and penalizes joint torque, mechanical power, loss of wheel-ground contacts, deviation from an upright orientation, and rapid action changes. Complete reward weights are given in Table~\ref{tab:rl_params}. The motor-level behaviors that execute the MPC commands include differential wheel torque for yaw, asymmetric leg extension for roll, and symmetric wheel torque for longitudinal acceleration. These responses emerge in the policy from the explicit command-tracking rewards and the symmetry, contact, and penalties incorporated in training. 

\subsection{Sim-to-Real Transfer}
The policy is trained and checked for functionality in simulation and deployed directly on hardware. Domain randomization is applied at the start of each training episode to vary ground friction coefficients, vehicle mass, CG offsets, and actuator gains over the ranges listed in Table~\ref{tab:rl_params}. During each episode, the robot is subjected to randomly timed external impulse forces and random velocity perturbations to improve disturbance rejection. The PD gains $K_{p,i}$ and $K_{d,i}$ are randomized during training to add robustness to gain uncertainty. Noise is injected at every step, and the previous action is included in the observation so the policy learns to compensate for the resulting actuator lag. Together, these strategies encourage feedback behaviors that generalize across conditions encountered during physical deployment and promote consistent sim-to-real transfer. 

\section{Experimental Setup}\label{sec:exp}

\begin{figure}[h]
    \centering
    \includegraphics[width=0.8\linewidth]{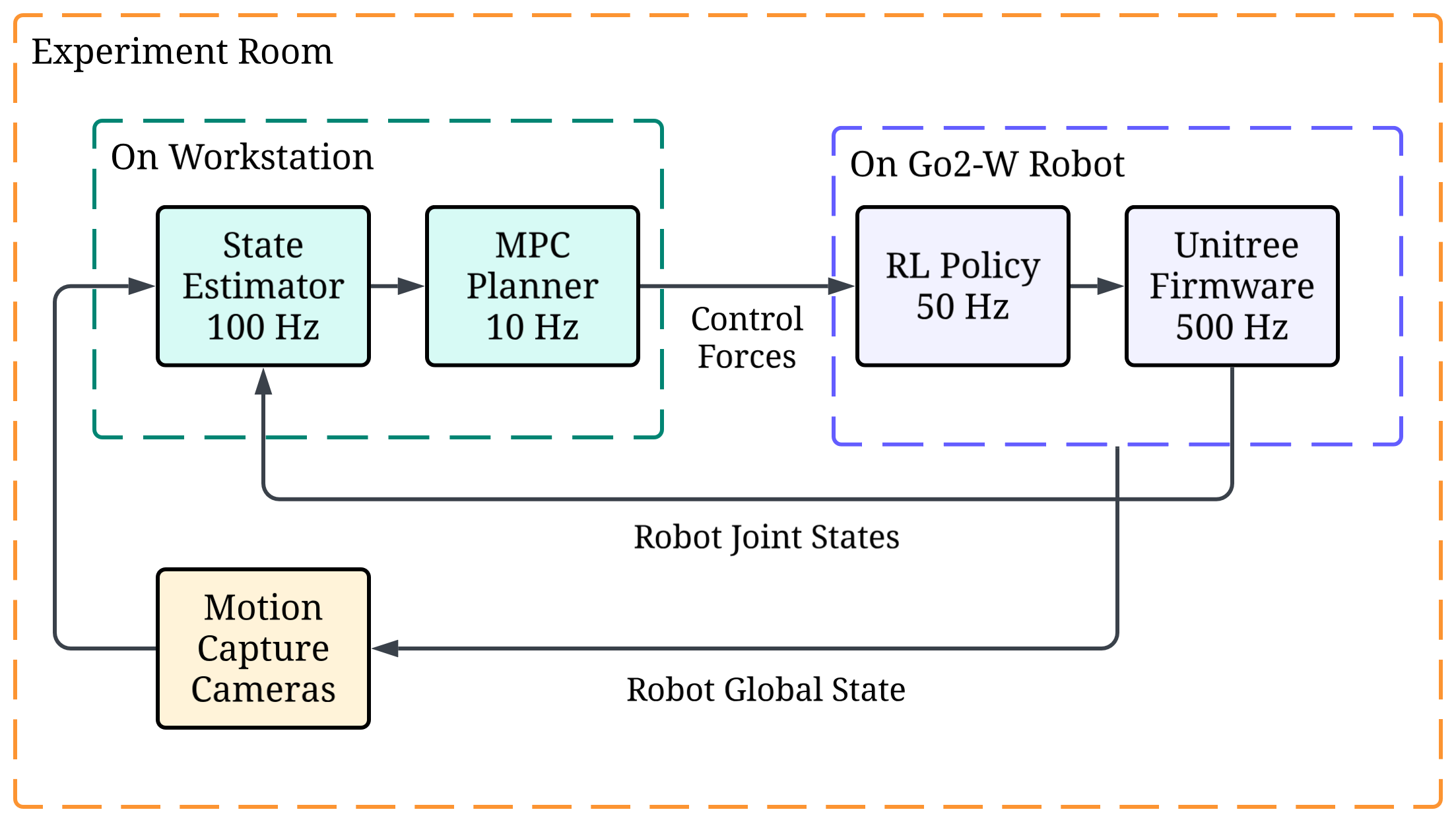}
    \caption{Experimental control and communication system.}
    \label{fig:exp}
\end{figure}

The experimental platform is the Unitree Go2-W quadruped robot (Fig. \ref{fig:go2w}). In its default operating mode the Go2-W combines wheeled driving with stepping, but for this application the control architecture constrains it to vehicle-style wheel-ground contact and an upright, symmetric driving stance with allowance for asymmetry only from active body lean through the leg joints. Experiments are conducted on a planar, indoor L-shaped track, with four turns, a $17.5$\,m centerline, and a $1.1$\,m width (Fig.~\ref{fig:raceline}). The combination of straight sections and corners of varying radii enables a comprehensive evaluation of the racing performance.

\begin{figure}[h]
    \centering
    \includegraphics[width=0.5\linewidth]{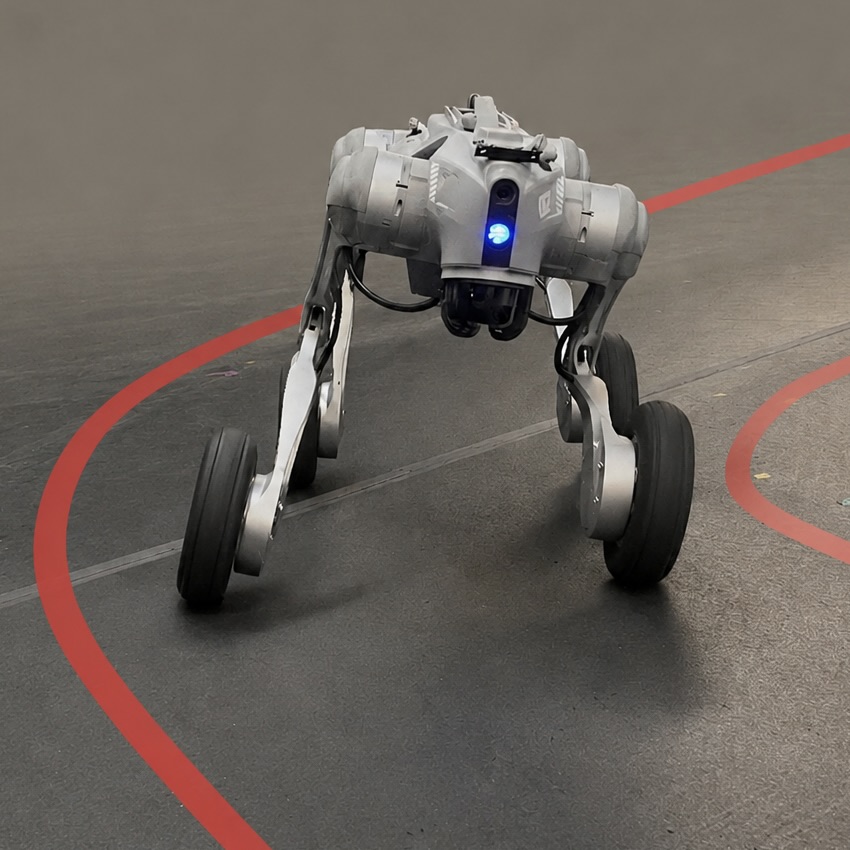}
    \caption{Unitree Go2-W applying active roll control during testing.}
    \label{fig:dog2}
\end{figure}
\subsection{Experimental Architecture}

The control architecture is illustrated in Fig.~\ref{fig:exp}. State estimation is provided by the OptiTrack Motion Capture system, which streams pose data at $100$\,Hz via WiFi and ROS\,2 to an off-board lab workstation (not physically connected to the robot). The workstation runs the MPC planner and publishes the resulting command sequence to the robot at $10$\,Hz. The MPC solves in a background thread while the $10$\,Hz timer publishes the last valid solution, even if the newest solve takes too long or is infeasible. The RL policy on-board the robot commands the robot movement at $50$\,Hz, and the Unitree firmware executes the commands at $500$\,Hz. 

\subsection{MPC Gain Tuning}\label{sec:tune}
The MPC dynamics in Sec.~\ref{sec:dyn} assume that the control inputs $[a_x, M_\psi, M_\theta]^\top$ are delivered ideally by the low-level layer. In practice, the delivered response differs from the command in magnitude and timing. We characterized this mismatch experimentally and used the result to inform the MPC cost weights $w$ in Eqn.~\eqref{eq:stagecost}. Based on data from experimental laps around the track, the control/response mismatch for the three control inputs was estimated by cross-correlation, and a single linear gain through the origin was fit on the delay-aligned signals by least squares. The identified delivery gains were approximately 0.95 for longitudinal acceleration, 0.76 for yaw moment (underdelivered), and 1.05 for roll moment (slightly over-delivered). The MPC cost weights were then hand tuned using these gains to obtain the values listed in Table~\ref{tab:params}. 

\section{Results}\label{sec:results}

\begin{figure}[h]
\centering
    \includegraphics[width=1.0\linewidth]{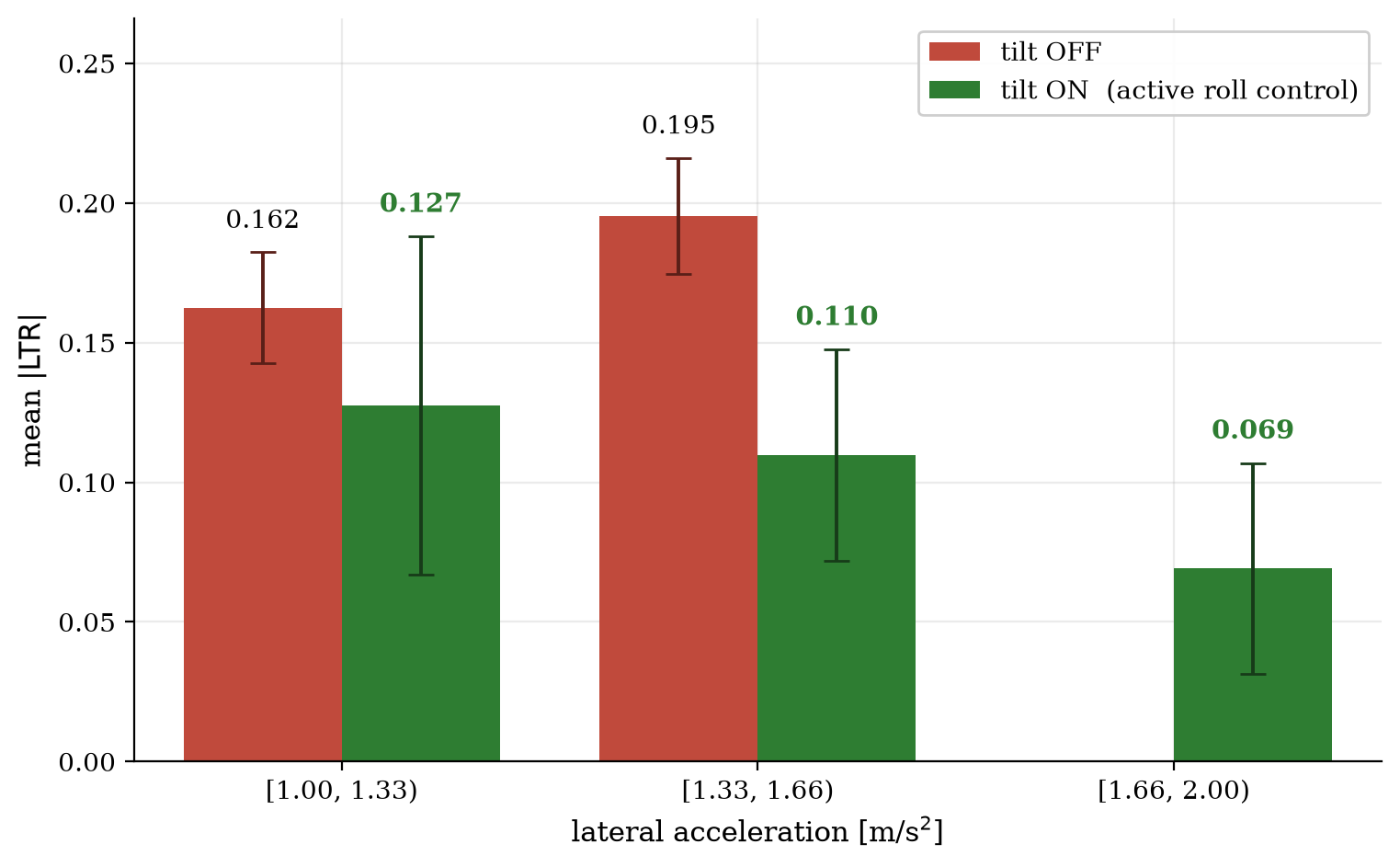}
    \caption{LTR vs. lateral acceleration for roll control vs. baseline.}
    \label{fig:plot}
\end{figure}

\label{tab:results}
\begin{table}[h]
\centering
\caption{Baseline (no roll control) vs. active roll control}
\label{tab:results}
\begin{tabular}{|l||c|c|c|}
\hline
\textbf{Metric} & \textbf{Roll OFF} & \textbf{Roll ON} & \textbf{$\Delta$ \%} \\ \hline
Mean lap time [s] & 11.03 & \textbf{10.32} & -6.4\% \\
Fastest lap [s] & 10.96 & \textbf{10.01} & -8.7\% \\
Slowest lap [s] & 11.27 & \textbf{10.82} & -4.0\% \\
Mean speed [m/s] & 1.46 & \textbf{1.47} & +1.1\% \\
Peak speed [m/s] & 2.11 & \textbf{2.23} & +5.7\% \\
Mean lateral accel. [m/s$^2$] & 0.794 & \textbf{0.844} & +6.3\% \\ 
Peak lateral accel. [m/s$^2$] & 1.64 & \textbf{1.98} & +21.3\% \\ 
Mean $|\mathrm{CTE}|$ [m] & 0.115 & \textbf{0.111} & -2.9\% \\
Peak $|\mathrm{CTE}|$ [m] & 0.492 & \textbf{0.438} & -11.1\% \\ \hline
\end{tabular}
\end{table}

Experimental results include 10 laps around the L-shaped track using the controller in Sec.~\ref{sec:exp} on the Go2-W with a maximum velocity of $3.0$\,m/s (enforced via Eqn.~\eqref{eq:ftocp_vbound}) in both the active roll (tilt ON) and baseline (tilt OFF) configurations. Both configurations share a raceline, planner, low-level policy, and hardware setup. The baseline's only difference is that the MPC commands $M_\theta=0$ throughout the experiment. The higher cornering capabilities of our proposed active roll controller yielded more aggressive tracking of the reference trajectory. This resulted in a peak lateral acceleration of $1.98$\,m/s$^2$ for tilt ON versus $1.64$\,m/s$^2$ for the baseline. At matched lateral acceleration during cornering, the active roll control reduced the mean LTR by up to $44\%$ (Fig.~\ref{fig:plot}). The active roll controller also outperformed the baseline in mean and peak lap time, speed, lateral acceleration, and cross-track error (CTE) (Table~\ref{tab:results}). The LTR had higher standard deviation in tilt ON vs tilt OFF (Fig.~\ref{fig:plot}). This distribution is non-Gaussian and multi-modal at corners, indicating systematic slight over-compensation of roll moment consistent with the roll over-delivery identified in Sec.~\ref{sec:tune}, and possibly suggesting an unmodeled communication delay from the MPC. The Go2-W leaning into a corner is shown in Fig.~\ref{fig:dog2}.

\section{Conclusion}\label{sec:conclusion}
We presented a hierarchical MPC and RL control framework for wheeled quadrupedal robots that incorporates active roll control to mitigate lateral load transfer during autonomous racing. We used this controller to demonstrate the effectiveness of active roll control on the wheeled quadruped robot on a track compared to a baseline controller that did not optimize roll angle. The results showed that the active roll control reduced mean LTR, improved lap time, and boosted peak lateral acceleration capability while maintaining stability beyond the range of the non-tilting baseline controller.

\section{Appendix}\label{sec:appendix}
\small
\tablecaption{Model and MPC parameter values. Reward weights have inverse units of their corresponding features.} 
\centering
\label{tab:params}
\begin{supertabular}{|c||c|c|c|}
\hline
\textbf{Symbol} & \textbf{Description} & \textbf{Value} & \textbf{Units} \\ \hline
\hline \multicolumn{4}{|c|}{\textit{Vehicle Physical Parameters}} \\ \hline
$m$              & Total vehicle mass            & 15.0          & kg          \\ 
$h$              & Center of gravity height      & 0.40          & m           \\ 
$w$              & Wheel track width             & 0.55          & m           \\ 
$L_f$            & c.g.\ to front axle distance  & 0.20          & m           \\ 
$L_r$            & c.g.\ to rear axle distance   & 0.30          & m           \\ 
$I_x$            & Roll inertia                  & 2.5           & kg$\cdot$m$^2$ \\ 
$I_z$            & Yaw inertia                   & 2.8           & kg$\cdot$m$^2$ \\ 
$C_{\alpha f}$   & Front cornering stiffness     & 800           & N/rad       \\ 
$C_{\alpha r}$   & Rear cornering stiffness      & 750           & N/rad       \\ 
$g$              & Gravitational acceleration    & 9.81          & m/s$^2$     \\ \hline
\multicolumn{4}{|c|}{\textit{MPC Solver Parameters}} \\ \hline
$N$              & Prediction horizon            & 35            & steps       \\ 
$\Delta t$       & Control period                & 0.10          & s           \\ \hline
\multicolumn{4}{|c|}{\textit{Hard Constraint Bounds}} \\\hline 
$v_{\min}$       & Minimum speed                 & 0.30          & m/s         \\ 
$v_{\max}$       & Maximum speed                 & 3.0           & m/s         \\ 
$a_{x,\min}$     & Min longitudinal acceleration & $-2.0$        & m/s$^2$     \\ 
$a_{x,\max}$     & Max longitudinal acceleration & 2.0           & m/s$^2$     \\ 
$M_{\theta,\max}$& Max roll moment magnitude     & 15.0          & N$\cdot$m   \\ \hline
\multicolumn{4}{|c|}{\textit{Soft Constraint Thresholds}} \\ \hline
$\beta_{\max}$        & Max sideslip angle (soft)     & 0.30          & rad         \\ 
$\dot\psi_{\max}$     & Max yaw rate (soft)           & $\pi/3$       & rad/s       \\ 
$\theta_{\max}$       & Max roll angle (soft)         & 0.50          & rad         \\ \hline
\multicolumn{4}{|c|}{\textit{Stage Cost Weights ($w$)}} \\ \hline
$w_\psi$              & Heading tracking weight                & 300           & --          \\ 
$w_v$                 & Speed tracking weight                  & 40            & --          \\ 
$w_p$                 & Position tracking weight               & 2000          & --          \\ 
$w_\beta$             & Sideslip soft-bound weight             & 200           & --          \\ 
$w_{\dot\psi}$        & Yaw rate soft-bound weight             & 2000          & --          \\ 
$w_{tb}$              & Track boundary penalty weight          & 16000         & --          \\ 
$w_{\theta^\star}$    & Optimal-roll (LTR) tracking weight     & 50            & --          \\ 
$w_\theta$            & Roll angle soft-bound weight           & 5000          & --          \\ \hline
\multicolumn{4}{|c|}{\textit{Control Regularization ($R = \mathrm{diag}(r_{a_x},\, r_{M_\psi},\, r_{M_\theta})$)}} \\ \hline
$r_{a_x}$        & Acceleration regularization   & 0.010         & --          \\ 
$r_{M_\psi}$     & Yaw moment regularization     & 0.001         & --          \\ 
$r_{M_\theta}$   & Roll moment regularization    & 0.001         & --          \\ \hline
\multicolumn{4}{|c|}{\textit{Rate Smoothing ($R_\Delta = \mathrm{diag}(r_{\Delta a_x},\, r_{\Delta M_\psi},\, r_{\Delta M_\theta})$)}} \\ \hline
$r_{\Delta a_x}$      & Acceleration rate weight  & 3.0           & --          \\ 
$r_{\Delta M_\psi}$   & Yaw moment rate weight    & 1.0           & --          \\ 
$r_{\Delta M_\theta}$ & Roll moment rate weight   & 1.0           & --          \\ \hline
\end{supertabular}

\centering
\tablecaption{RL low-level policy parameters.}
\label{tab:rl_params}
\begin{supertabular}{|c|c|c|} \hline
\textbf{Description} & \textbf{Value} & \textbf{Units} \\ \hline \hline
\multicolumn{3}{|c|}{\textit{Policy Network Architecture}} \\ \hline
Hidden layer widths (actor \& critic) & 512, 256, 128    & --     \\ 
Parallel training environments         & 4096            & --     \\ \hline
\multicolumn{3}{|c|}{\textit{Observation Noise}} \\ \hline
Angular velocity noise                 & $\pm 0.2$       & rad/s  \\
Linear velocity noise                  & $\pm 0.1$       & m/s    \\
Joint position noise                   & $\pm 0.01$      & rad    \\
Joint velocity noise                   & $\pm 1.5$       & rad/s  \\ \hline
\multicolumn{3}{|c|}{\textit{Reward Weights}} \\ \hline
Longitudinal acceleration tracking               & $+3.0$          & --     \\
Yaw moment tracking                      & $+1.5$          & --     \\
Roll moment tracking                      & $+1.5$          & --     \\
Upright orientation                    & $+1.0$          & --     \\ \hline
Vertical base velocity                 & $-2.0$          & --     \\
Roll/pitch angular rate                & $-0.05$         & --     \\
Leg joint torques (L2)                 & $-2.5\times10^{-5}$ & --  \\
Leg joint power                        & $-2.0\times10^{-5}$ & --  \\
Diagonal leg symmetry (FR$\leftrightarrow$RL, FL$\leftrightarrow$RR) & $-0.05$ & -- \\
Action rate (L2)                       & $-0.01$         & --     \\
Non-foot body contact                  & $-12.0$         & --     \\
Foot contact force magnitude           & $-1.5\times10^{-4}$ & --  \\ \hline
\multicolumn{3}{|c|}{\textit{Domain Randomization}} \\ \hline
Static friction coefficient            & $\mathcal{U}(0.3,\,1.0)$  & --   \\
Dynamic friction coefficient           & $\mathcal{U}(0.3,\,0.8)$  & --   \\
Base mass additive offset              & $\mathcal{U}(-1.0,\,3.0)$ & kg   \\
Other body mass scale                  & $\mathcal{U}(0.7,\,1.3)$  & --   \\
Center of mass offset (per axis)       & $\mathcal{U}(\pm 0.05)$   & m    \\
Actuator stiffness scale               & $\mathcal{U}(0.5,\,2.0)$  & --   \\
Actuator damping scale                 & $\mathcal{U}(0.5,\,2.0)$  & --   \\
External impulse force at reset        & $\mathcal{U}(\pm 10)$     & N    \\
External impulse torque at reset       & $\mathcal{U}(\pm 10)$     & N$\cdot$m \\
Random push velocity (interval)        & $\mathcal{U}(\pm 0.5)$    & m/s  \\
Push interval                          & $\mathcal{U}(10,\,15)$    & s    \\ \hline
\end{supertabular}
\end{document}